\documentclass[letterpaper, 10 pt, conference]{ieeeconf}  

\IEEEoverridecommandlockouts                              

\usepackage{amsmath} 
\usepackage{cuted}
\usepackage{graphicx}
\usepackage{subcaption}
\usepackage{balance}

\usepackage{eso-pic}
\usepackage{hyperref}

\AddToShipoutPictureBG*{%
  \AtPageUpperLeft{%
    \hspace{0.5\paperwidth}%
    \raisebox{-1.2cm}{%
      \makebox[0pt][c]{%
        \footnotesize
        \begin{tabular}{c}
          Accepted to 2025 IEEE International Conference on Robotics and Automation (ICRA), Atlanta, USA \\
          DOI: \href{https://doi.org/10.1109/ICRA55743.2025.11128781}{10.1109/ICRA55743.2025.11128781}
        \end{tabular}
      }%
    }%
  }%
  \AtPageLowerLeft{%
    \hspace{0.5\paperwidth}%
    \raisebox{1.0cm}{%
      \makebox[0pt][c]{%
        \parbox{0.85\paperwidth}{%
          \centering
          \scriptsize
          \copyright\ 2025 IEEE. Personal use of this material is permitted. Permission from IEEE must be obtained for all other uses, in any current or future media, including reprinting/republishing this material for advertising or promotional purposes, creating new collective works, for resale or redistribution to servers or lists, or reuse of any copyrighted component of this work in other works.
        }%
      }%
    }%
  }%
}

\title{\LARGE \bf
Improved Bag-of-Words Image Retrieval\\ with Geometric Constraints for Ground Texture Localization
}

\author{Aaron Wilhelm$^{1}$ and Nils Napp$^{1}$
\thanks{$^{1}$School of Electrical and Computer Engineering,
        Cornell University, Ithaca, NY 14853, USA. {\tt\small ajw344@cornell.edu, nnapp@cornell.edu}}
        }

\begin{document}

\maketitle
\thispagestyle{empty}
\pagestyle{empty}

\begin{abstract}
Ground texture localization using a downward-facing camera offers a low-cost, high-precision localization solution that is robust to dynamic environments and requires no environmental modification.
We present a significantly improved bag-of-words (BoW) image retrieval system for ground texture localization, achieving substantially higher accuracy for global localization and higher precision and recall for loop closure detection in SLAM.
Our approach leverages an approximate $k$-means (AKM) vocabulary with soft assignment, and exploits the consistent orientation and constant scale constraints inherent to ground texture localization.
Identifying the different needs of global localization vs. loop closure detection for SLAM, we present both high-accuracy and high-speed versions of our algorithm.
We test the effect of each of our proposed improvements through an ablation study and demonstrate our method's effectiveness for both global localization and loop closure detection.
With numerous ground texture localization systems already using BoW, our method can readily replace other generic BoW systems in their pipeline and immediately improve their results.
\end{abstract}

\section{Introduction} 




Localization is a critical ability for mobile robots to perform a wide variety of tasks~\cite{Cadena2016}.
This work focuses on localization using ground texture, captured by a downward-facing camera, which offers a low-cost, high-accuracy solution without requiring any environment modification~\cite{hdground}.
Unlike LiDAR or outward-facing cameras, ground texture localization offers robustness to dynamic environments, occlusions, and lighting changes because the camera can be shielded.
Additionally, the close proximity to the ground enables a high level of localization precision.
This robustness and precision, along with its other advantages, has led to research exploring ground texture localization in autonomous vehicles~\cite{Kozak2016, Fang2007, Li2021}, factory settings~\cite{Fukase2016, Sola2021}, and warehouse environments~\cite{Kelly2000, Gadd2015, Xu2024}.

Recently, several ground texture localization methods have been proposed that use bag-of-words (BoW) for visual place recognition (VPR) for loop closure or global localization~\cite{hdground, Chen2018, hart2023monocular, Wang2024}.
Loop closure and, more generally, relocalization within a global reference frame is important for mobile robots to reduce drift caused by compounding odometry errors~\cite{Cadena2016}.
In the case of global localization, VPR systems such as BoW need high precision to localize correctly.
However, often BoW image retrieval is the first step in a loop closure pipeline with subsequent filtering processes, and as such necessitates high recall to avoid missing potential valid loop closures~\cite{Garg2021}.
As demonstrated later in this work though, directly applying BoW to ground texture images without considering geometric constraints can lead to suboptimal performance, likely due to the limited and potentially less distinct image domain.

\begin{figure}[tbp]
    \centering
    \vspace*{1mm}\includegraphics[width=0.95\linewidth]{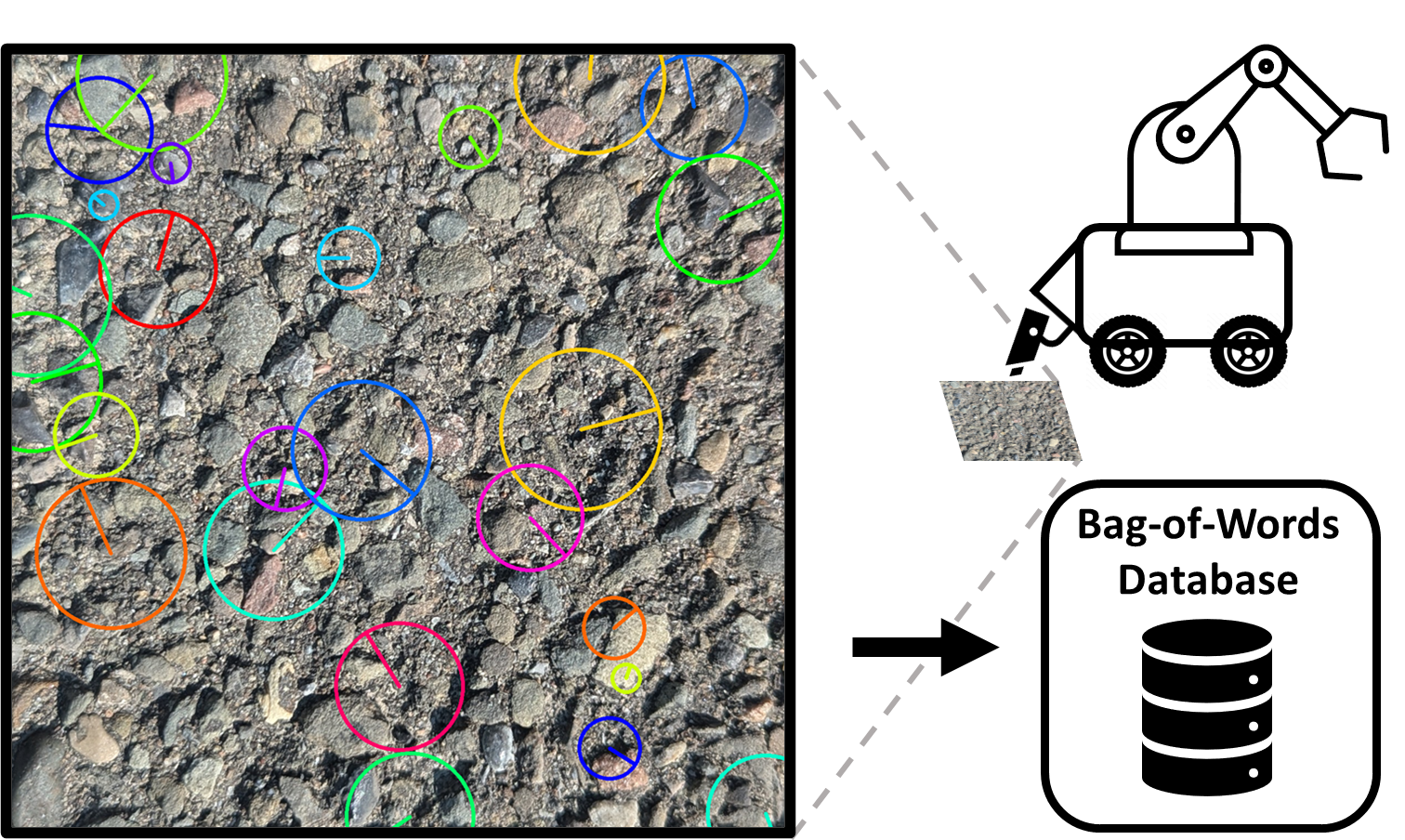}
    \caption{Our algorithm converts images of the ground texture into a bag-of-words representation to localize a mobile robot.}
    \label{fig:paper_overview}
\end{figure}

Recognizing the importance of BoW to high-accuracy localization and its already prevalent use for ground texture localization, here we present several techniques to substantially improve BoW for this application.
We leverage extensive previous work on BoW regarding approximate $k$-means (AKM) vocabularies and soft assignment, where descriptors can be assigned to several visual words, to greatly improve the effectiveness of the visual vocabulary.
Furthermore, we exploit the consistent orientation and constant scale constraints inherent to ground texture localization to improve both precision and recall.
Acknowledging the differing requirements of global localization and loop closure in SLAM, we propose two tailored versions of our algorithm: a high-accuracy version (localization frequency $>5~\text{Hz}$) for global localization and a higher-speed version ($>13~\text{Hz}$) optimized for loop closure detection for SLAM systems.
Notably, this work can be immediately applied to previous ground texture localization works that use BoW to directly improve their accuracy and overall results.
To facilitate this, we have released our code on our lab website.

\subsection{Related Work} 

\subsubsection{Ground Texture Localization}

While researchers have thoroughly investigated using ground texture for visual odometry~\cite{Lovegrove2011, Rajesh2024}, utilizing the ground to globally localize or detect loop closures is a relatively underexplored area.
Initially, several different methods such as template matching~\cite{Kelly2000} and edge detection~\cite{Fang2009} were employed for global localization using the ground.
Over time though, approaches that localized via feature matching with prebuilt maps became popular, but these required a prior pose estimate to reduce the search space~\cite{Gadd2015, Kozak2016}.
StreetMap was the first system to use BoW for ground texture global localization and did not need a prior pose, but SLAM was not shown~\cite{Chen2018}.

Zhang et al. soon after introduced MicroGPS, which matched features from a query ground image with those in a database, spatially filtering outliers to achieve high-precision localization~\cite{microgps}.
More recently, our L-GROUT algorithm~\cite{Wilhelm2024} built upon MicroGPS, demonstrating state-of-the-art accuracy and introducing a lightweight version using ORB features~\cite{Rublee2011} for faster feature extraction and localization.
While fast and accurate, the approximate nearest neighbor (ANN) search structures used in MicroGPS and L-GROUT require extensive computation time to build, resulting in a necessary mapping stage before localization.
Hart et al.~\cite{hart2023monocular} presented one of the first ground texture SLAM systems, using feature-based visual odometry and BoW with multiple filtering steps for loop closure and drift correction.
Finally, while BoW-based methods are the predominant approach for image retrieval in ground texture localization, Radhakrishnan~\cite{Radhakrishnan2021} explored deep metric learning, which achieved limited success compared to established techniques~\cite{Wilhelm2024}.
Given BoW's prevalence in ground texture localization, we focus our efforts on specializing BoW for this specific domain.

\subsubsection{Visual Bag-of-Words}
Visual BoW, which we will refer to as BoW for the rest of this paper, is an image retrieval method where image features are extracted from a query image and then matched to the closest visual word in a pre-trained vocabulary to build a BoW vector~\cite{Sivic2003}.
This BoW vector is then queried against an inverse index to efficiently retrieve the most similar database images.
While initially framed as a vector similarity problem, image retrieval using BoW can be reinterpreted as a weighted voting system where descriptors are assigned to words and then vote for images containing those words~\cite{Jegou2010}.

Originally, $k$-means clustering was used to train vocabularies~\cite{Sivic2003}, but its poor scaling due to long training times led to the introduction of the hierarchical $k$-means (HKM) vocabulary~\cite{Nister2006}.
HKM hierarchically clusters training descriptors, enabling faster vocabulary construction and word assignment.
However, quantization at each level of the HKM tree can result in significant word assignment errors, especially with noisy descriptors~\cite{Philbin2007}.
This limitation was addressed with the approximate $k$-means (AKM) vocabulary~\cite{Philbin2007}, which uses a pre-built ANN search structure for efficient descriptor assignment.
While AKM reduces quantization error, its hard assignment of descriptors to single words remains a source of error.
Soft assignment was therefore introduced to improve robustness by assigning descriptors to multiple visual words~\cite{Philbin2008}.

Several techniques have been developed to enhance BoW image retrieval.
Spatial verification, also known as geometric matching, ensures geometric consistency between matched images and the query image.
This can be achieved by verifying the existence of a consistent transform between the query and database images~\cite{Philbin2007}, or by applying a weak geometric constraint at the keypoint level~\cite{Jegou2010}, which is the approach we adopt.
Beyond geometric constraints, research has also explored incorporating additional feature information, such as color~\cite{Vigo2010} or scale~\cite{Khan2014}, into the BoW framework.

Many popular loop closure detection systems have used BoW.
For example, FAB-MAP 2.0~\cite{Cummins2011} employs a probabilistic BoW approach that utilizes an AKM vocabulary in its appearance-based SLAM algorithm for large-scale place recognition.
In contrast, DBoW~\cite{Galvez2012} implements BoW with an HKM vocabulary and fast-to-compute binary descriptors.
It further refines results by checking candidate loop closures for temporal and geometric consistency to mitigate false positives.
Because DBoW is a high-performing, widely-used system and has already been applied to ground texture SLAM~\cite{hart2023monocular}, it serves as an ideal baseline for our specialized BoW approach.

\section{System Description}

\begin{figure}[tbp]
    \centering
    \vspace*{2mm}\includegraphics[width=0.99\linewidth]{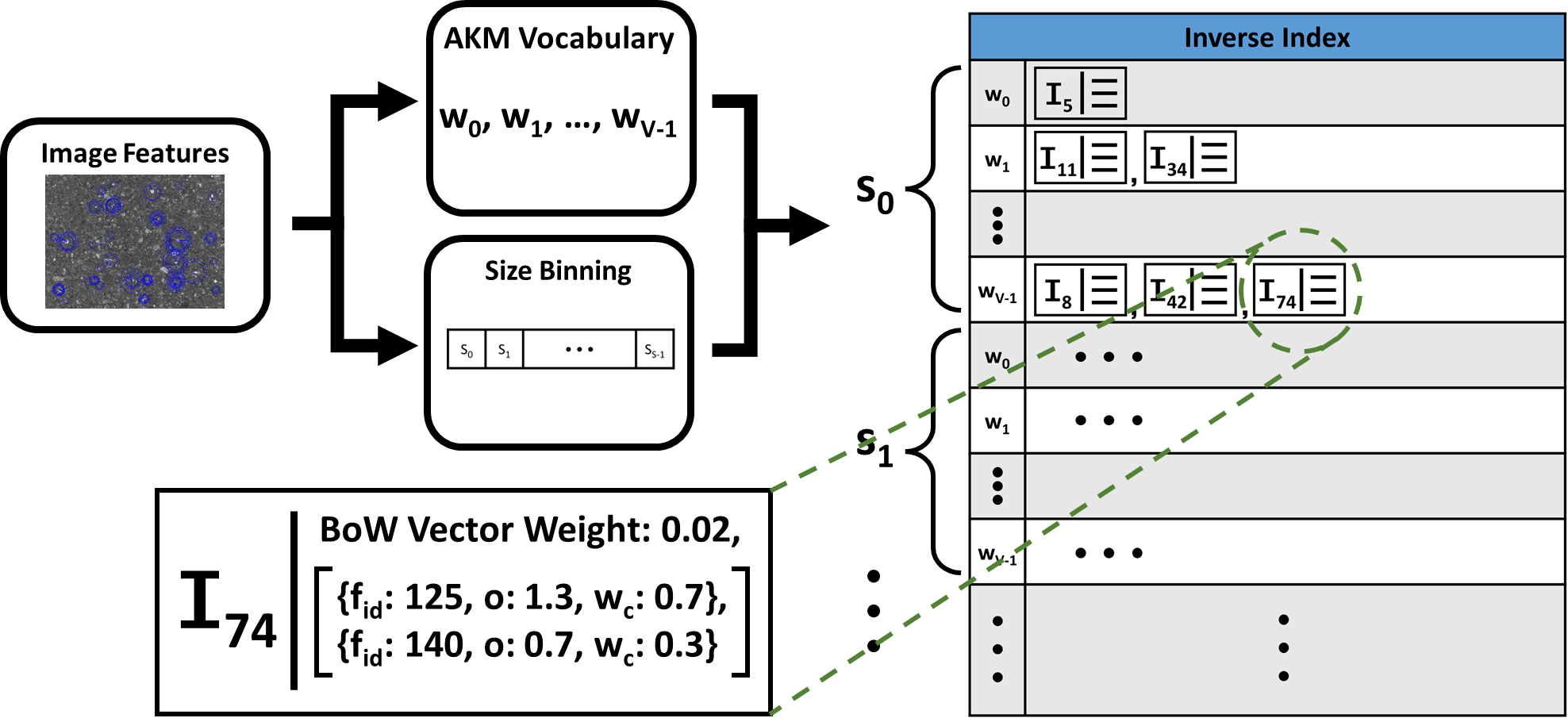}
    \caption{Our proposed BoW method. Image features are extracted from the ground image and each descriptor is assigned to a visual word via an AKM vocabulary. Additionally, the size of each feature is binned into one of $S$ pre-trained bins. Then, when inserting entries into the inverse index, each element is inserted into the row that corresponds to that feature's vocabulary word and size. Each entry consists of an image ID, the word's weight in the BoW vector, and a list of features that correspond to the word in that image. We store each feature's ID, orientation, and contributing weight to later use for orientation verification.}
    \label{fig:system_diagram}
\end{figure}


Fig.~\ref{fig:system_diagram} shows our modifications to the standard BoW method to optimize it for ground texture localization.
Two of our modifications, quantizing features based upon keypoint size and performing orientation verification, are derived from the geometric problem constraints of ground texture localization.
The other two modifications, the use of an AKM vocabulary and soft assignment, are techniques from previous research in BoW that have yet to be applied to ground texture localization.
In the following subsections we explain each of these changes in more detail.

\subsection{Binning Features By Size}
    The constant camera height in ground texture localization causes keypoints corresponding to the same physical feature to have identical or near identical sizes~\cite{microgps}.
    Algorithms like MicroGPS and L-GROUT exploited this constraint by binning keypoints by size, improving query speed and accuracy.
    Inspired by this, we integrate this technique into our BoW approach by binning features by size and concatenating their BoW vectors together to create the image's BoW vector, similar to the work of Khan et al.~\cite{Khan2014}.

    Our method begins by training a standard visual vocabulary of size $V$ using all descriptors from the training images.
    We also partition the features into $S$ distinct size bins.
    These bins are determined either by size percentile or, for feature extractors with discrete scales like ORB, by pyramid level.
    The thresholds defining these bins are stored for subsequent feature assignment.

    
    To convert an image into a BoW vector for database addition or querying, we first assign each detected keypoint to one of the $S$ size bins and its descriptor to a visual word.
    BoW vectors are then computed separately for each size bin and concatenated together.
    Finally, this concatenated BoW vector is $L_2$-normalized and either added to the database or used for querying.
    
    This process treats descriptors with the same visual word but different sizes as distinct features.
    In practice, this effectively means that we expand our original vocabulary size to $S \times V$, increasing our inverse index by a factor of $S$.
    This greatly increases our accuracy and speeds up database lookups since the BoW vectors are more sparse, leading to fewer collisions in the inverse index.

    As shown later in Section~\ref{sec:system_characterization}, dividing the descriptors into $S$ sizes before visual word assignment is substantially more effective than simply increasing the visual vocabulary size by a factor of $S$.
    This superior performance stems from the inherent geometric constraint of consistent keypoint size in ground texture localization, which is more reliable and discriminative than further refinement of the visual words.

\subsection{Consistent Orientation Verification}
    The constrained camera motion inherent to ground texture localization results in a key geometric property: correctly matched feature pairs across images will have similar differences between their orientations.
    Our prior work L-GROUT capitalized on this by filtering image origin votes based on orientation consistency~\cite{Wilhelm2024}.
    We adapt this principle to BoW, applying weak geometric verification by grouping votes from features with similar orientation differences.

    We implement this by augmenting the inverse index to store the orientation of each feature matched to a word.
    At query time, for each potential matching image we accumulate that image's score over $R$ bins, where each bin holds the votes for a distinct quantized orientation difference.
    The final image score is the max score of its $R$ bins and only the database keypoints corresponding to the max orientation bin are returned.


    We note that while many BoW loop closure systems employ a geometric check after image retrieval as a step to filter out false positive loop closures, the key distinction of our method is that we integrate an initial geometric check into the inverse index.
    This enables us to filter out incorrect matches during image retrieval, significantly improving accuracy, albeit with a slight increase in query time as demonstrated in Section~\ref{sec:system_characterization}.

\subsection{Approximate $k$-Means Vocabulary}
    Although hierarchical $k$-means (HKM) vocabularies are common in BoW-based SLAM systems due to their efficiency, they are known to suffer from significant quantization errors, reducing retrieval accuracy~\cite{Philbin2008}.
    To mitigate this, we utilize an approximate $k$-means (AKM) vocabulary, which, despite its slower descriptor assignment compared to HKM, provides significantly higher-quality word assignments.


    Furthermore, the improved word assignments allow for direct feature matching between query and database images based on identical visual word and size bin assignments.
    Consequently, it does not require the nearest neighbor distance ratio test to verify feature matches, unlike DBoW~\cite{Galvez2012} or the ground texture SLAM method by Hart et al.~\cite{hart2023monocular}.
    This streamlined matching process accelerates the feature matching process and reduces memory requirements by eliminating the need to store ORB descriptors or maintain a direct index.

\subsection{Soft Assignment}
    When quantizing descriptors to visual words, hard assignment can cause boundary descriptors to be allocated to different visual words with the introduction of minimal noise~\cite{Philbin2008}.
    This is problematic because these inconsistently assigned descriptors have the same weight in the BoW vector similarity calculation as consistently assigned descriptors.
    To address this, we implement soft assignment and assign each descriptor to its $r$ closest visual words.
    Each assignment receives a weight of $\exp\left(-\frac{d^2}{2\sigma^2}\right)$, where $d$ is the descriptor's distance to the visual word and $\sigma$ is an empirically determined spatial scale.
    These weights are then $L_1$-normalized across the $r$ assignments for each keypoint, reflecting assignment confidence.
    The resulting normalized weights are used for BoW vector similarity voting and, if localization is performed, for RANSAC.

    Soft assignment greatly increases the accuracy of image retrieval, but since there are more vocabulary words per image, the constructed BoW vectors are less sparse.
    While this has minimal impact on the time to insert a BoW vector into the inverse index, it does increase the query time since the number of potential matches for a query grows approximately linearly as $r$ increases~\cite{Philbin2008}.
    Therefore, we propose a high-accuracy version of our algorithm that uses a soft assignment ($r=3$) and, for applications such as SLAM where query time is critical, we present a high-speed version without soft assignment.

\subsection{Data Structures}
To implement our proposed changes, we make several modifications to the traditional inverse index as illustrated in Fig.~\ref{fig:system_diagram}.
First, since we divide features by size into $S$ bins, the inverse index requires $S$ times more rows, one for each combination of visual word and size bin.
We modify each entry in a row of the inverse index to contain an image ID, the weight of the word in the BoW vector, and a list of the keypoint information from the database image that corresponds to this visual word.
This list contains each keypoint's ID, orientation, and contributing weight, which is calculated as the keypoint's soft assignment weight divided by the total soft assignment weights of all image keypoints assigned to this word.
At query time, the contributing weights allow us to allocate the original weight of the word in the BoW vector into the correct orientation difference bins of the matching image if orientation discrepancies between the keypoints exist.
The keypoint ID is stored so later the keypoint's (x, y) coordinate can be retrieved for pose estimation during localization.



As previously mentioned, our method does not require storing a direct index or ORB feature descriptors.
As a result, the high-speed version of our algorithm without soft assignment requires less memory per image than DBoW, resulting in improved scalability.
However, the high-accuracy version using soft assignment leads to denser BoW vectors and requires storing contributing weights, thus slightly increasing memory usage per database image.




\section{Experiments}
\label{sec:experiments}

We evaluate our algorithm against the image retrieval component of DBoW, a widely used system that has been integrated into a ground texture SLAM pipeline~\cite{hart2023monocular} and represents a typical BoW-based approach.
Our evaluation focuses on the core BoW image retrieval process; therefore, we do not incorporate temporal or spatial filtering techniques commonly used in SLAM loop closure systems.
We begin with a system characterization in Section~\ref{sec:system_characterization} where we perform an ablation study, timing experiments, and a scaling analysis.
We then present tests showcasing our algorithm's global localization performance in Section~\ref{sec:global_localization} and its ability to detect loop closures in Section~\ref{sec:loop_closure_detection}.

Unless otherwise specified, all of the experiments have the following parameters.
We use the HD Ground dataset~\cite{hdground} and for training, database, and query images we extract 250 ORB features from each image.
ORB features are chosen for their speed and prevalence in SLAM systems, and this also facilitates a more straightforward comparison to DBoW, which relies on ORB features.
However, our proposed methods are applicable to continuous descriptors such as SIFT~\cite{Lowe2004} as well.
While vocabularies are trained on the provided training sets for the main four textures, we use the asphalt dataset for experiments since it is the largest dataset (32,251 images covering an area of $106~\text{m}^2$) and relevant to applications such as autonomous vehicles, industrial robots, and delivery robots.
Although not shown, our results are not texture specific and generalize to other textures within the HD Ground dataset.

By default, the DBoW algorithm uses an HKM vocabulary.
We target a vocabulary size of $V=100,000$ and train a vocabulary with a branching factor $k=10$ and $L=5$, resulting in a vocabulary of 92,055 words since branching stops when a node has too few descriptors.
Since we can set the size of the AKM vocabulary directly, we are able to train a vocabulary with size $V=100,000$ for our algorithms.
While a slight variation in vocabulary sizes exists due to their distinct training processes, we later demonstrate that our algorithm's accuracy improvements significantly outweigh the effect of the minor increase in vocabulary size.
All vocabularies use the standard \textit{tf-idf} weighting when creating BoW vectors~\cite{Sivic2003}.
For our high-accuracy version of the algorithm we use an AKM vocabulary with $S=8$ size bins (one for each ORB pyramid level), $R=6$ orientation bins, and soft assignment with $r=3$ and $\sigma=580$.
Our high-speed algorithm version has the same parameters but does not do soft assignment, setting $r=1$.
All timing tests were performed on a computer with a 13th Gen Intel\textsuperscript{\textregistered} Core\textsuperscript{\texttrademark} i7-13700 processor.

\subsection{System Characterization}
\label{sec:system_characterization}

\subsubsection{Ablation Study}
    To evaluate the individual impact of our proposed modifications, we conduct an ablation study.
    We use DBoW as a baseline and incrementally add our modifications: size binning, the AKM vocabulary, soft assignment, and orientation verification.
    Performance is measured using mean average precision (mAP), calculated by averaging the average precision of the top 1000 returned images for each query image in the asphalt dataset's test set.
    A returned image is considered a true positive if it overlaps with the query image by at least 25\%, ensuring sufficient visual similarity.

    The results in Table~\ref{table:ablation_study} clearly demonstrate that each modification substantially improves mAP on the asphalt dataset.
    Soft assignment provides the largest performance gain, followed by size binning, orientation verification, and the AKM vocabulary.
    Because these techniques are largely independent, their combined effect is substantial, as shown by the high mAP values achieved by both our high-accuracy and high-speed algorithm versions.
    In contrast, the baseline DBoW performs poorly, highlighting the challenges of image retrieval without these optimizations.

\begin{table}[t]
\centering
\vspace*{1.5mm}
\caption{An ablation study of our algorithm. DBoW is the baseline and size binning (SB), the AKM vocabulary (AKM), soft assignment (SA), and orientation verification (OV) are added. The highest mAP is in bold.}
\addtolength{\tabcolsep}{-2pt}
\begin{tabular}{|c| c|}
 \hline
 \textbf{Algorithm} & \textbf{mAP} \\ [0.5ex] 
 \hline\hline
 DBoW & 0.026\\ 
 \hline
DBoW + SB & 0.091\\
 \hline
DBoW + AKM & 0.073\\
 \hline
DBoW + AKM + SA ($r=3$) & 0.191\\
\hline
DBoW + OV & 0.083\\
\hline
Ours Fast (BoW + SB + AKM + OV) & 0.258\\
\hline
Ours (BoW + SB + AKM + SA ($r=3$) + OV) & \textbf{0.559}\\
\hline
\end{tabular}
\label{table:ablation_study}
\end{table}


While size binning with $S$ bins increases the inverse index size by a factor of $S$, this approach is significantly more effective than simply increasing the vocabulary size by the same factor.
To illustrate this, we used an HKM vocabulary ($k=10$, $L=4$) with $S=8$ size bins, resulting in a vocabulary roughly 10 times smaller than the default.
Despite this size reduction, the size-binned vocabulary achieved nearly four times the mAP (0.100 vs. 0.026), clearly demonstrating the benefits of size binning.

\subsubsection{General Timing}
Next, we measured the computation times for key stages of DBoW and our algorithms.
Table~\ref{table:speed_tests} reports the mean and standard deviation of these timings, measured over 100 randomly selected asphalt images.
To simulate realistic operating conditions, these tests were performed using databases pre-populated with the 32,251 images from the asphalt dataset.
Although each algorithm uses ORB, we present the feature extraction times as a reference point for the other parts of the algorithms.

DBoW requires 6.7 ms to transform ORB features into BoW vectors, which is faster than the 24 ms needed by our algorithms.
This difference is attributed to DBoW's use of an HKM vocabulary, in contrast to our use of an AKM vocabulary.
DBoW is also the fastest for database insertion, but given that all times are below 2 ms, this difference is negligible for overall runtime.
Our high-speed algorithm version without soft assignment achieves the fastest query time since the size binning makes the BoW vectors larger and consequently results in a sparser inverse index.
Meanwhile, our high-accuracy algorithm with soft assignment is the slowest since the additional word assignments make the inverse index more dense.
However, with a query frequency exceeding $5~\text{Hz}$, we feel that this tradeoff for the significant accuracy improvement shown in the previous section is justifiable.

\begin{table*}[t]
\centering
\vspace*{1mm}
\caption{Timing comparison of DBoW, our high-accuracy algorithm, and our high-speed algorithm. The database (DB) insertion and query times exclude the time to transform the descriptors into the BoW vector. The mean and standard deviation are presented, with the fastest times in bold.}
\renewcommand{\arraystretch}{1.1} 
\addtolength{\tabcolsep}{-2pt}
\begin{tabular}{|c|| c| c| c| c| c|}
 \hline
 \textbf{Algorithm} & \textbf{Feature Extraction Time (ms)} & \textbf{BoW Transform Time (ms)} & \textbf{DB Insertion Time (ms)} & \textbf{DB Query Time (ms)}\\ [0.5ex] 
 \hline\hline
 DBoW & $19.1 \pm 2.5$ & $\mathbf{6.7 \pm 0.0}$ & $\mathbf{0.0 \pm 0.0}$ & $50 \pm 5.5$\\ 
 \hline
 Ours (High Acc.) & $18.9 \pm 2.5$ & $24 \pm 0.3$ & $1.4 \pm 0.5$ & $149.6 \pm 14.9$\\
 \hline
  Ours (Fast) & $18.9 \pm 2.5$ & $23.1 \pm 0.3$ & $0.6 \pm 0.4$ & $\mathbf{31.9 \pm 3.7}$\\
 \hline
\end{tabular}
\label{table:speed_tests}
\end{table*}

\subsubsection{Scaling Time Tests}

\begin{figure}[tbp]
    \centering
    \vspace*{2mm}\includegraphics[width=0.95\linewidth]{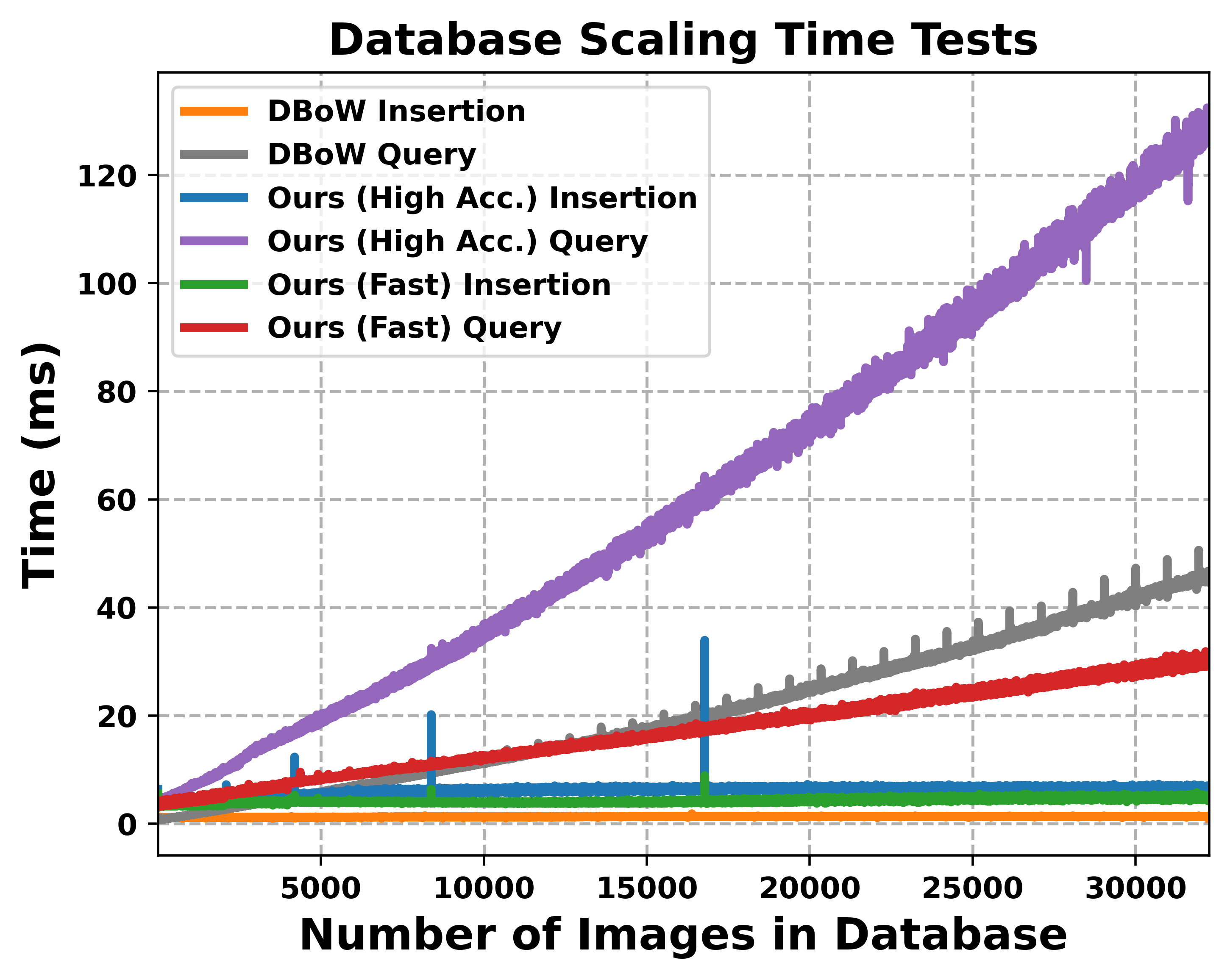}
    \caption{The database insertion and querying timings as the size of the database increases.}
    \label{fig:asphalt_scaling_speed_test}
\end{figure}

To evaluate scaling performance in a SLAM context, we measured database insertion and query times as the database size increased, as shown in Fig.~\ref{fig:asphalt_scaling_speed_test}.
All algorithms exhibit near-constant database insertion times, regardless of database size.
Initially, our high-speed algorithm has slightly slower query times than DBoW due to the overhead of orientation verification.
However, its sparser inverse index from size binning leads to better scaling as the database grows.
Our high-accuracy algorithm with soft assignment, in contrast, exhibits a query time that scales approximately three times slower than DBoW due to the denser inverse index.

\subsection{Global Localization}
\label{sec:global_localization}

\begin{figure}[tbp]
    \centering
    \vspace*{2mm}\includegraphics[width=0.95\linewidth]{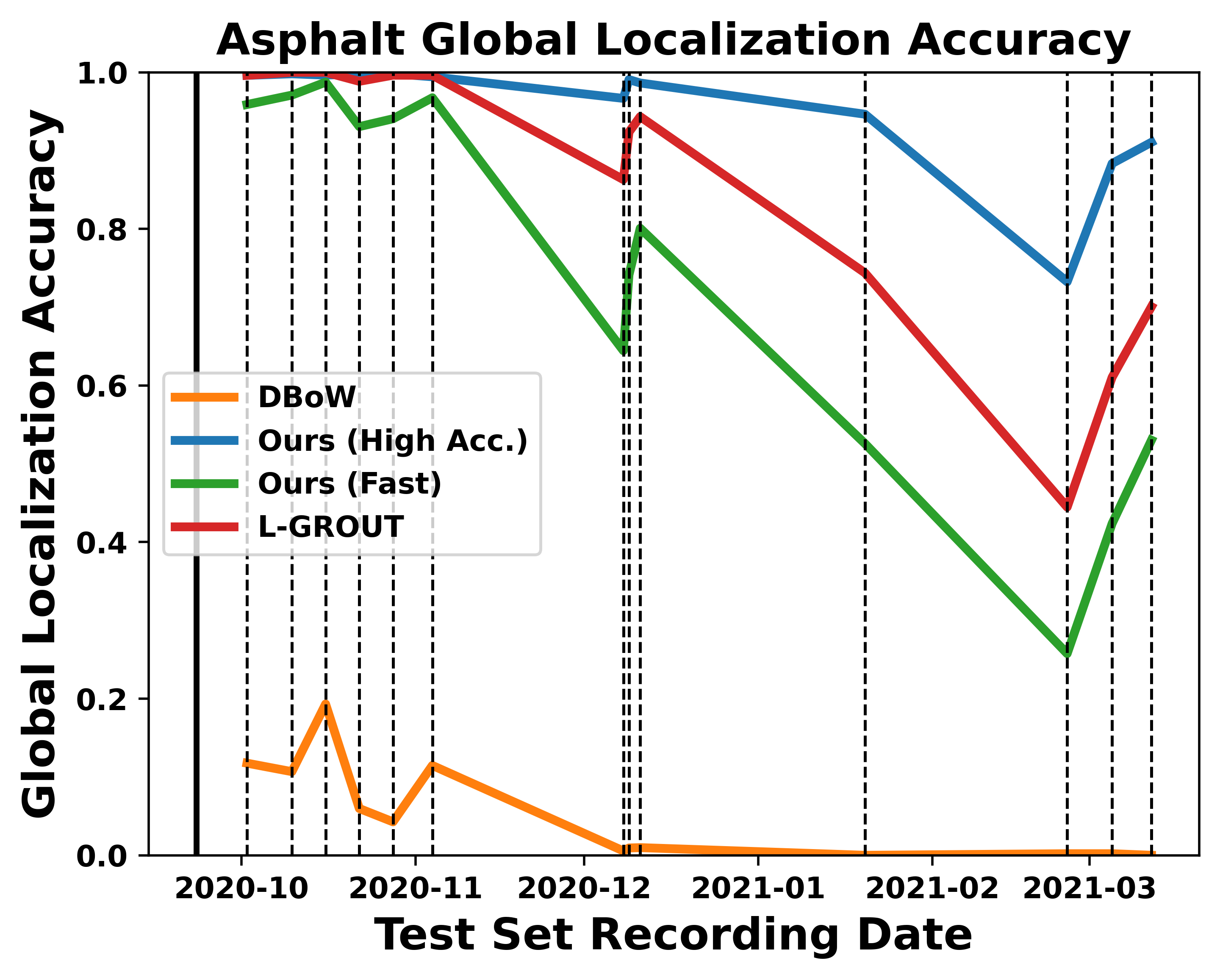}
    \caption{Global localization performance of different algorithms on the HD Ground asphalt dataset over time~\cite{hdground}. The bold vertical line indicates the database recording date, the dashed lines indicate test set recording dates.}
    \label{fig:asphalt_global_loc}
\end{figure}

Additionally, we tested the global localization performance of the algorithms using the asphalt test sets collected over approximately six months.
For each test set, we report the percentage of images correctly localized.
Following previous testing standards for HD Ground, an image is considered correctly localized if its estimated position is within 4.8 mm translation and 1.5 degrees rotation of the ground truth.
Localization with DBoW is performed by feature matching using its direct index and the nearest neighbor distance ratio, followed by RANSAC to estimate the transformation, as described in the original publication~\cite{Galvez2012}.
Our algorithms match features assigned to the same visual word and size bin, and then apply RANSAC to estimate the final pose.
A weighted RANSAC method is used when soft assignment is enabled.

For comparison, we include results from our previous state-of-the-art ground texture localization algorithm, L-GROUT~\cite{Wilhelm2024}.
L-GROUT uses locality-preserving projections to reduce feature dimensionality to 16 dimensions and incorporates size binning with 8 size-based bins.
Its projections were trained across all textures in the HD Ground dataset.
During map construction, 250 features are extracted per image, with the 10 largest features plus 40 randomly selected features added to the database.
For test images, all 250 extracted features are used for querying.

Fig.~\ref{fig:asphalt_global_loc} shows that our high-accuracy algorithm outperforms L-GROUT.
This is especially notable because L-GROUT requires an offline mapping stage, while our BoW algorithm builds its database online.
Initially, our high-speed algorithm performs comparably to both L-GROUT and our high-accuracy algorithm, but its performance degrades over time.
This degradation is likely due to environmental noise (e.g., dirt, scuff marks) altering or obscuring the originally mapped descriptors and introducing new ones.
Soft assignment, used in our high-accuracy algorithm, naturally handles these variations better than the hard assignment used in our high-speed algorithm and DBoW.
Finally, DBoW exhibits significantly lower localization performance throughout the testing period and towards the end it struggles to localize at all.
This underscores the impact of our improvements to BoW for ground texture global localization.

\subsection{Loop Closure Detection}
\label{sec:loop_closure_detection}

Finally, we evaluate the loop closure detection performance of our algorithms.
As a BoW query is often the first step in a loop closure pipeline before additional filtering, a useful evaluation metric is Recall@$N$.
Recall@$N$ measures the percentage of queries that have at least one correct result when the top $N$ images are returned~\cite{Torii2015}.
A correct match is defined as a returned image that overlaps the query image by at least 25\%.
The Recall@$N$ provides insight into the algorithm's ability to retrieve any true positives before subsequent filtering steps.

\begin{figure}[tbp]
  \centering
  \begin{subfigure}[b]{0.49\linewidth}
    \includegraphics[width=\linewidth]{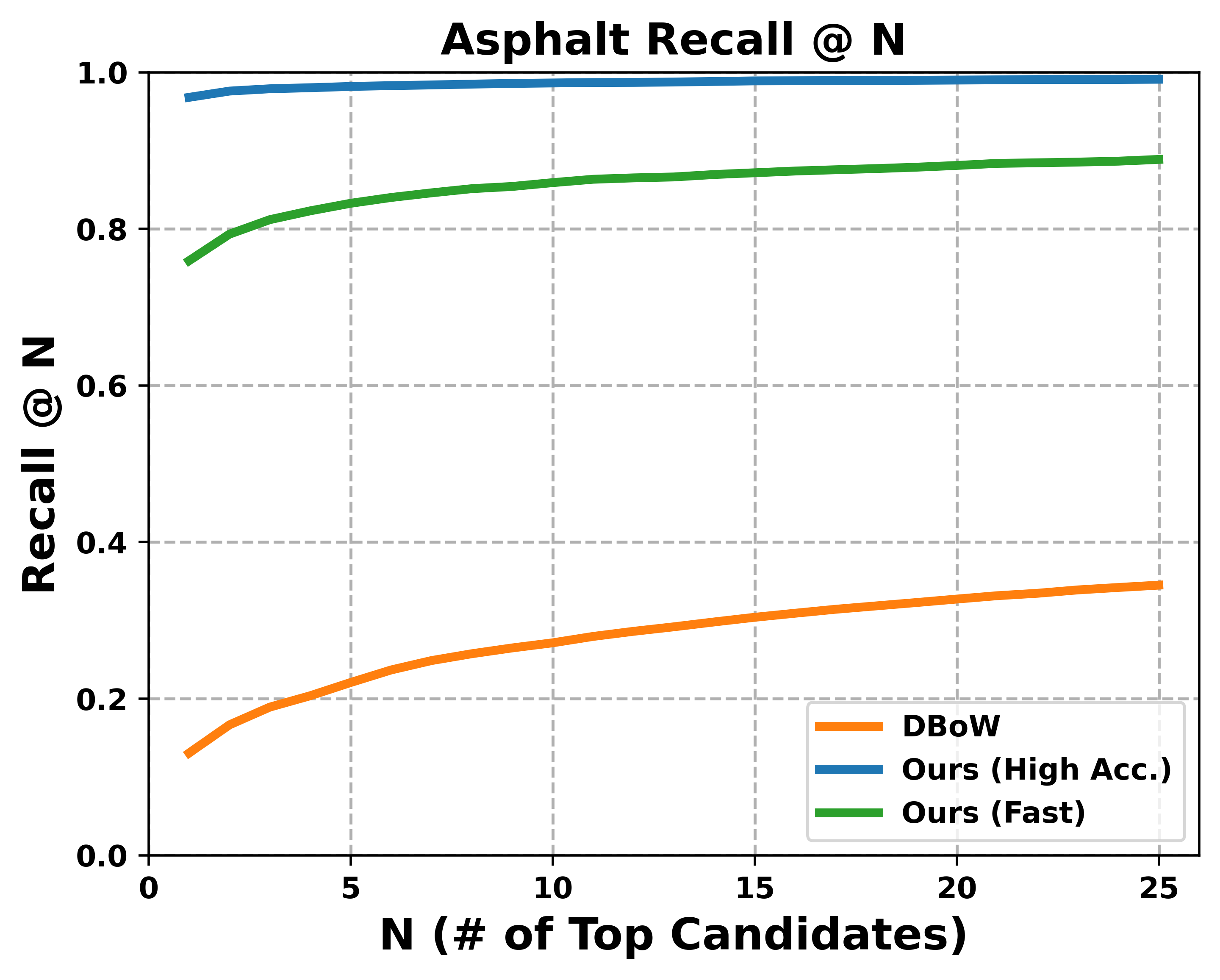}
  \end{subfigure}
  \begin{subfigure}[b]{0.49\linewidth}
    \includegraphics[width=\linewidth]{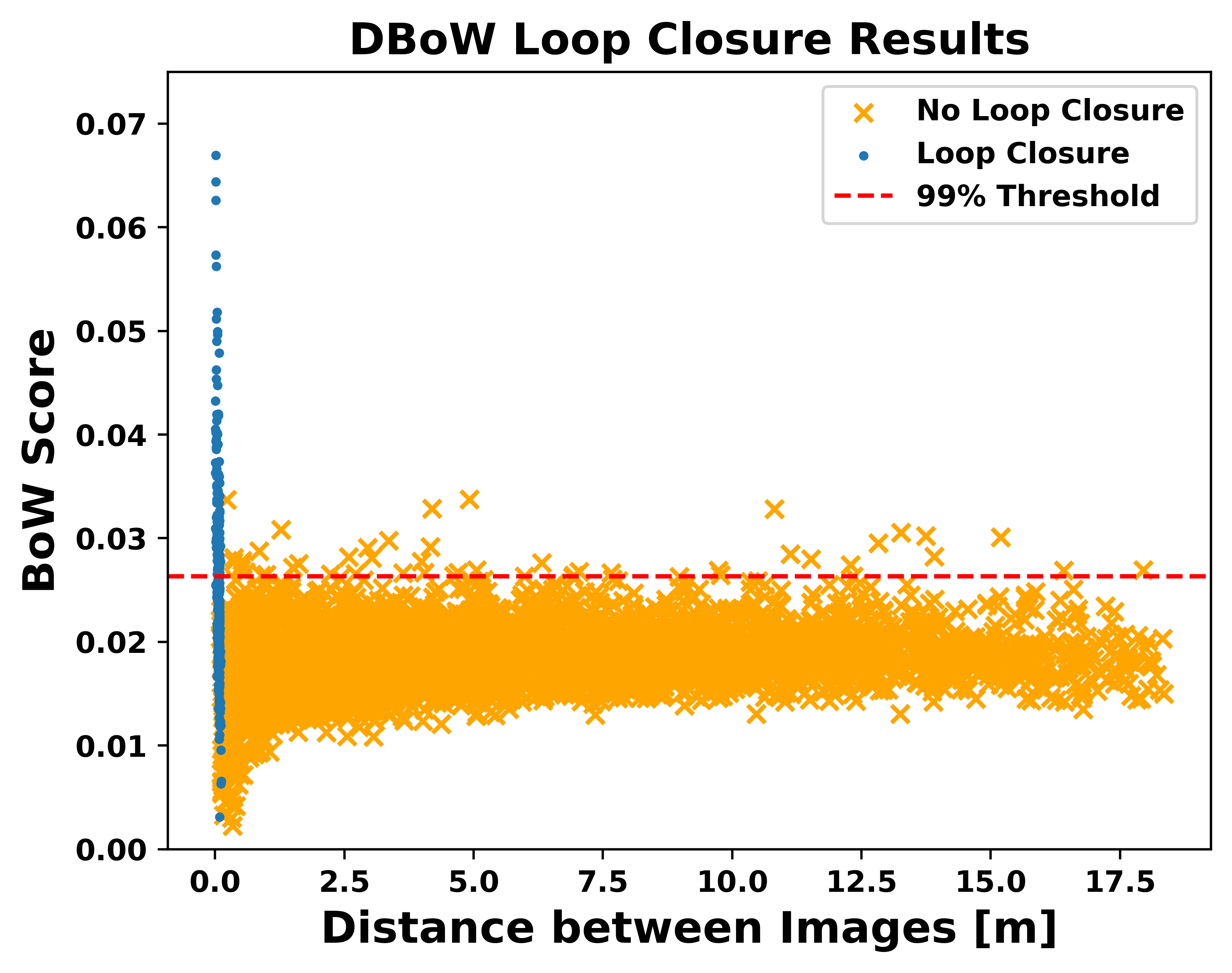}
  \end{subfigure}
  \begin{subfigure}[b]{0.49\linewidth}
    \includegraphics[width=\linewidth]{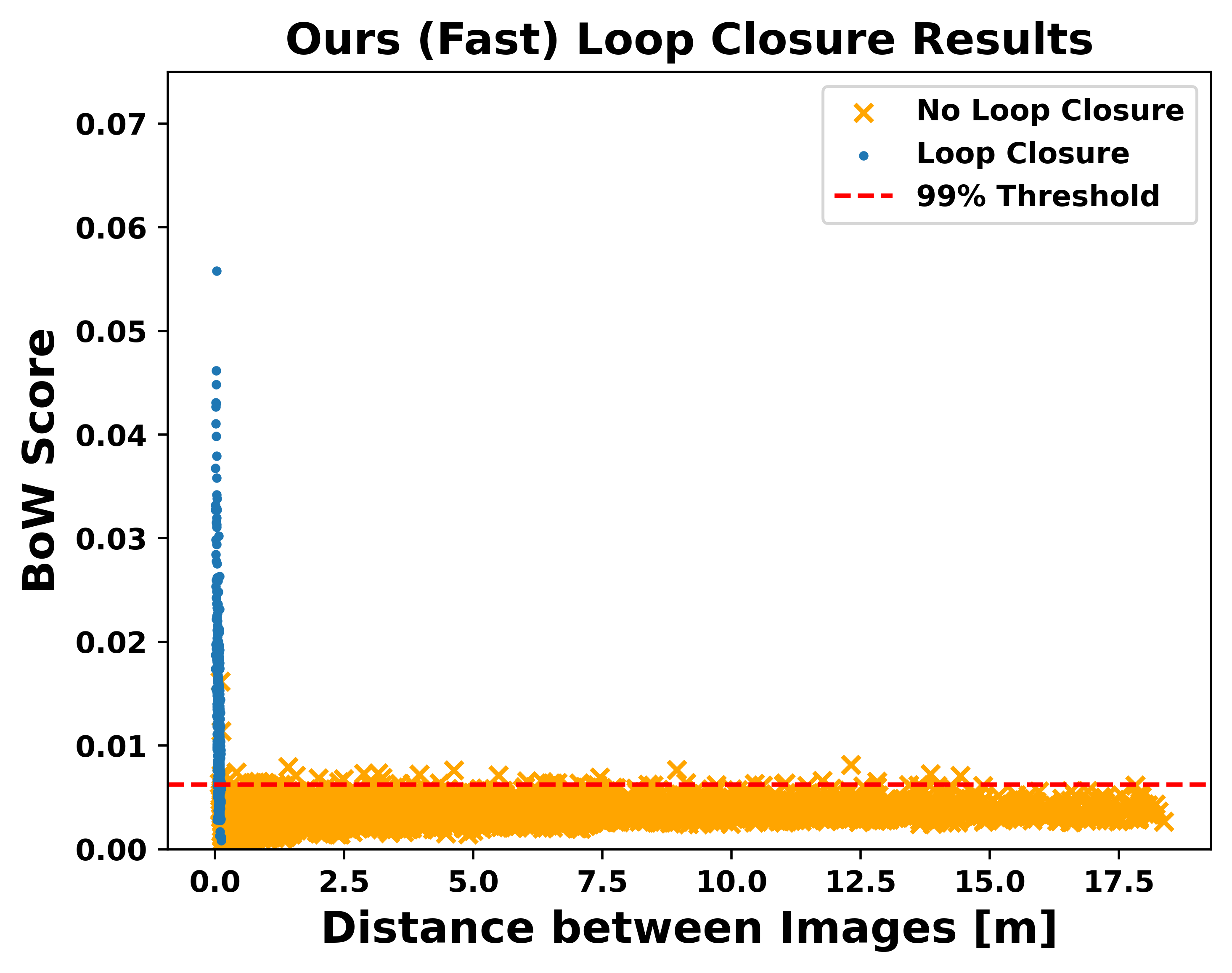}
  \end{subfigure}
    \begin{subfigure}[b]{0.49\linewidth}
    \includegraphics[width=\linewidth]{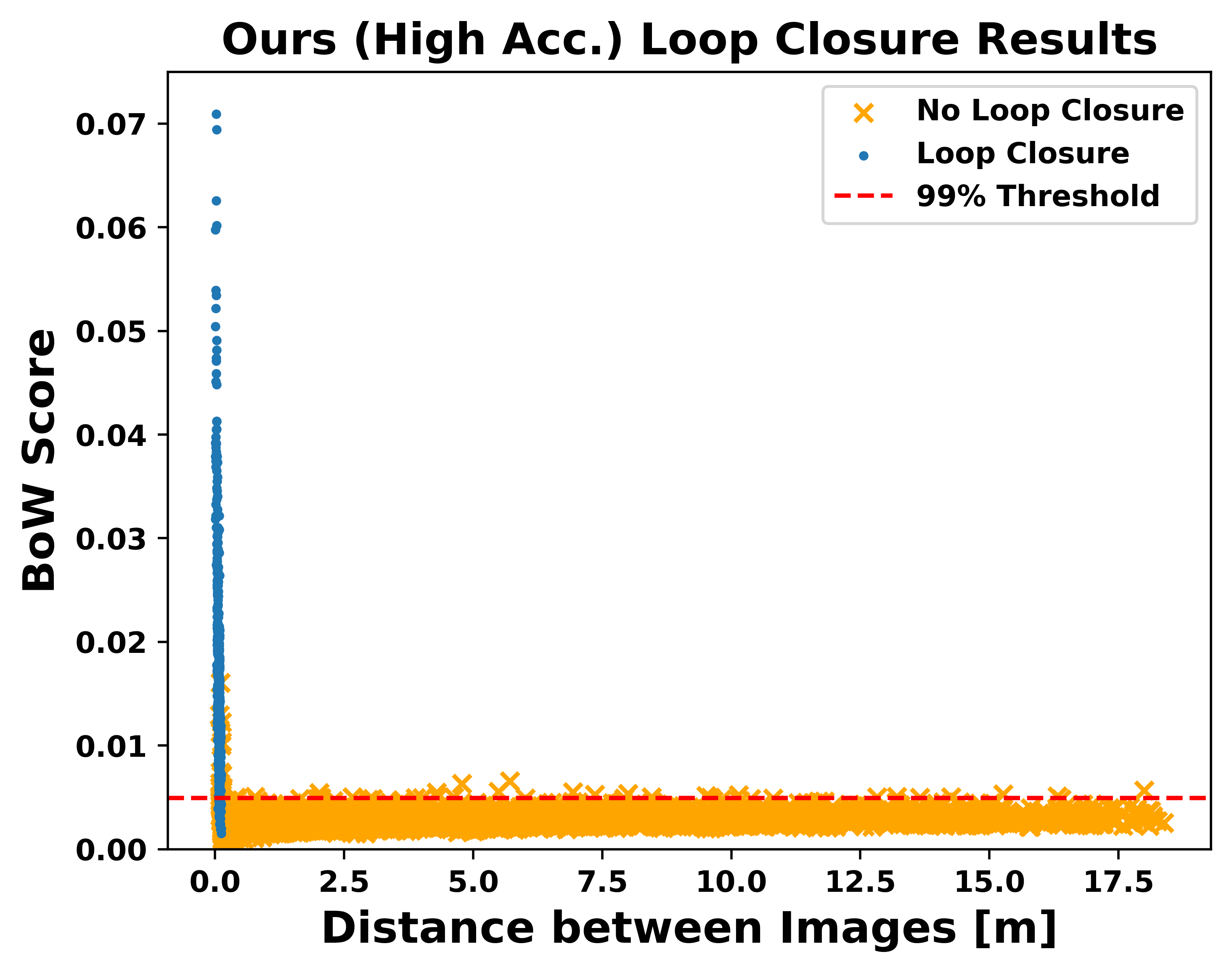}
  \end{subfigure}
    \caption{
    Top Left: The Recall@$N$ for DBoW and our algorithms on the asphalt dataset. Other: BoW similarity scores for potential loop closures for DBoW and our algorithms.}
    \label{fig:loop_closure_results}
\end{figure}

Fig.~\ref{fig:loop_closure_results} (Top Left) plots the Recall@$N$ for DBoW and our algorithms across all asphalt test set images.
Both of our algorithms significantly outperform DBoW in Recall@$N$, with our high-accuracy algorithm achieving near-perfect Recall@$N$ and demonstrating exceptional loop closure detection capability.

While Recall@$N$ indicates an algorithm's ability to retrieve potential loop closures, it does not reflect the distinguishability of true positives from false positives.
Typically, SLAM systems using BoW employ a filtering step, rejecting images with BoW similarity scores below a defined threshold since they are unlikely to be true loop closures.
Ideally, a large difference exists between the scores of true and false closures to improve loop closure recall and consequently localization accuracy.

In Fig.~\ref{fig:loop_closure_results} we graph the BoW similarity scores for the top ten retrieved images from one of the asphalt test sets.
We also select a hypothetical threshold value to reject 99\% of the false positives (in practice, a SLAM system designer would select their own threshold value to achieve their desired precision and recall).
These plots reveal a greater separation between true and false positive score distributions for our algorithms compared to DBoW, indicating improved discriminative ability.
Using these hypothetical thresholds, DBoW would retain only 32.5\% of true loop closures, while our high-speed and high-accuracy algorithms would retain 71.01\% and 91.21\%, respectively.
These findings demonstrate the superior ability of our algorithms to identify true loop closures while rejecting false positives.

\section{Conclusion} 
We have presented several modifications to the standard BoW algorithm that significantly enhance its performance for both global localization and loop closure detection in ground texture applications.
These modifications include size binning, orientation verification, AKM vocabulary usage, and soft assignment.
We introduce a high-accuracy version, leveraging soft assignment, that achieves state-of-the-art global localization at over $5~\text{Hz}$, and a high-speed version (without soft assignment) that provides robust localization at over $13~\text{Hz}$.
Both versions are readily applicable to existing BoW-based localization and SLAM systems for ground textures, and their implementations are available on our lab website.
Future work will focus on refining other aspects of the SLAM pipeline to further exploit the geometric constraints of ground texture localization, ultimately improving the overall SLAM capabilities of these systems.




\balance
\bibliography{IEEEabrv, refs}

\end{document}